\documentclass{article}

\usepackage[final]{nips_2016}

\usepackage[utf8]{inputenc} 
\usepackage[T1]{fontenc}    
\usepackage{hyperref}       
\usepackage{url}            
\usepackage{booktabs}       
\usepackage{amsfonts}       
\usepackage{nicefrac}       
\usepackage{microtype}      
\usepackage{graphicx}
\usepackage{subcaption}
\usepackage{caption}

\title{Capsule Network Performance on Complex Data}

\author{
  Edgar Xi \\
  Machine Learning Department\\
  Carnegie Mellon University\\
  Pittsburgh, PA 15213 \\
  \texttt{edgarxi@cmu.edu} \\
  \And
  Selina Bing\space \space \space \space Yang Jin \\
  Computer Science Department \\
  Carnegie Mellon University \\
  Pittsburgh, PA 15213 \\
  \texttt{\{zbing, yangjin\}@andrew.cmu.edu}
}

\begin{document}

\maketitle

\section{Introduction}
In recent years, convolutional neural networks (CNN) have played an important role in the field of deep learning. Variants of CNN's have proven to be very successful in classification tasks across different domains. However, there are two big drawbacks to CNN's: their failure to take into account of important spatial hierarchies between features, and their lack of rotational invariance [1]. As long as certain key features of an object are present in the test data, CNN's classify the test data as the object, disregarding features' relative spatial orientation to each other. This causes false positives. The lack of rotational invariance in CNN's would cause the network to incorrectly assign the object another label, causing false negatives. To address this concern, Hinton et al. propose a novel type of neural network using the concept of capsules in a recent paper. With the use of dynamic routing and reconstruction regularization, the capsule network model would be both rotation invariant and spatially aware. [1] \\ \\ 
The capsule network has shown its potential by achieving a state-of-the-art result of 0.25\% test error on MNIST without data augmentation such as rotation and scaling, better than the previous baseline of 0.39\%. To further test out the application of capsule networks on data with higher dimensionality, we attempt to find the best set of configurations that yield the optimal test error on CIFAR10 dataset.

\section{Overview of Capsule Networks}
Capsule networks represent a recent breakthrough in neural network architectures. They achieve state of the art accuracy on the MNIST dataset, a feat achieved traditionally by deep convolutional neural network architectures. Capsule networks introduce an alternative to translational invariance other than pooling through the use of modules, or capsules. Two key features distinguish them from CNN's: layer-based squashing and dynamic routing. Whereas CNN's have their individual neuron's squashed through nonlinearities, capsule networks have their output squashed as an entire vector. Capsules replace the scalar-output feature detectors of CNNs with vector-output capsules and max-pooling with routing-by-agreement. Capsnet architectures typically include several convolution layers, with a capsule layer in the final layer. 

\subsection{Dynamic Routing}
Capsules output a vector, which means that it is possible to selectively choose which parent in the layer above the capsule is sent to. For each potential parent, the capsule network can increase or decrease the connection strength. This routing by agreement is much more effective at adding invariance than the primitive routing introduced by max-pooling. 

\subsection{Reconstruction Regularization}
Whereas traditional CNN's prevent overfitting by using dropout, Capsule networks are regularized with a reconstruction autoencoder. During training, all activity vectors are masked except for the activity vector corresponding to the correct digit. This activity vector is then used to reconstruct the input image. The output of the digit is then used to compute the loss. This encourages the network to learn a more general representation of images. 

\subsection{Related Work}

A paradigm shift in the field of Machine Learning occurred when Geoffrey Hinton, Ilya Sutskever, and Alex Krizhevsky from the University of Toronto created a deep convolutional neural network architecture called AlexNet[2]. The architecture they created beat state of the art results by an enormous 10.8\% on the ImageNet challenge. 

An incremental advance in  that attempted to address the representational limitations of CNN's came from Hinton et al in 2011,[3] in which the concept of "capsules" first entered machine learning terminology. Capsules addressed the representational inefficiencies of CNN with transformation matrices, allowing networks to automatically learn part-whole relationships and thus generalize to novel viewpoints, something traditional CNN's exhibit exponentially poor performance on. 

Introduced by Hinton et al., Capsule Networks represent a similar paradigm shift in the field of Deep Learning[1], with a novel learning architecture, but their classification ability on datasets more complex than MNIST has yet to be seen.

The idea of Dynamic Routing - a crucial concept in Capsule Networks, has a biological basis, as demonstrated by Hinton et al.[7]. The importance of dynamic routing procedure is highlighted through a biologically plausible model, in which dynamic connections and object-based frames of reference "are used to generate shape
descriptions that can be used for object recognition"[7].

\section{Methodology}
We start with a baseline model, Hinton's MNIST model with 3 color channels. We explore the effect of a variety of model modifications, ranging from stacking more capsule layers to trying out different parameters.

\begin{itemize}
\item \textbf{Stacking more capsule layers}: The baseline model architecture is built specifically for MNIST, a relatively low-dimensionality dataset. To account for the complex underlying features of CIFAR10, stacking more capsule layers to account for the increasingly intricate relationship among features may improve the representational power of the network.
\item \textbf{Increasing the number of primary capsules}: As the dimensionality of data increases, a typical approach of traditional neural network is to stack more layers and increase the number of hidden units. Since each capsule represents a small group of neurons, we believe that increasing the number of primary capsules would yield a better accuracy as richer features may be learned.
\item \textbf{Ensemble averaging}: In ensemble averaging, a group of networks are trained together, and their predictions are averaged at test time. This prompts different networks to learn distinct features and typically yields a higher accuracy than one network can achieve by itself. Hinton's paper on capsule networks yielded a 10.6\% test error with an ensemble of 7 models. [1]
\item \textbf{Modify scaling of reconstruction loss}: Due to the difference in data complexity between MNIST and CIFAR10, the scaling of each pixel for reconstruction loss may be different as well. We experiment with the scaling factor and analyze scaling factor's influence on overfitting and rate of convergence.
\item \textbf{Increase number of convolution layers before capsule layer}: The higher dimensionality of CIFAR10 data entails a more complex encoding of the image. The hypothesis is that creating a more complex image encoding before feeding it into the capsule layer may yield higher accuracy.
\item \textbf{Customized activation function}: We use our customized activation function $f(x) = (1 - \frac{1}{e^{|x|}})\frac{x}{|x|}$ in place of the squash function. We hypothesize that our activation function will be sensitive to small changes in x, and thus lead to greater separation of classes. For example $f(x)$ already converges to 1 near the value $x=3$.
\item \textbf{Utilize none of the above category}: Capsule Networks have the tendency to explain everything in the image, so accuracy could be improved by adding a none of the above category as the 11th category.
\end{itemize}

\section{Datasets}
To explore capsule network's performance on data with higher dimensionality, we use the CIFAR-10 dataset, which is a subset of 80 million tiny images collected by Alex Krizhevsky, Vinod Nair, and Geoffrey Hinton.[5] The dataset consists of 32x32 colored and labeled images coming from 10 different classes, in which each class contains exactly 6,000 images: \textit{airplane, automobile, bird, cat, deer, dog, frog, horse, ship, truck}. The dataset is, by design, completely mutually exclusive: for instance, there is no overlap between \textit{automobile} and \textit{truck}. 50,000 images will be used as training data, and 10,000 images will be used as validation data. The CIFAR-10 dataset is chosen because it is an established computer vision dataset with an abundance of existing classification accuracy reports, and the images are sufficiently complex compared to MNIST, enabling us test on data with higher dimensionality.

Since capsule networks were introduced only a few weeks ago as of the writing of this paper, there have not been a lot of results on its performance on CIFAR10. The best test error achieved on CIFAR10 with capsule network is trained with 3 routing iterations on 24 $\times$ 24 patches of the image, and 64 different types of primary capsules, as stated in Hinton et al's original paper. [1] The state-of-the-art validation error on the CIFAR10 dataset is 4.50\% using fractional max-pooling within a convolutional neural network. [4] Based on the stellar performance of capsule network on the MNIST dataset, we believe given a good configuration, we can close the gap between the current test error and the best test error for CIFAR10. 

\section{Results}
After experimenting with different combinations of approaches mentioned in the \texttt{methods} section, we choose the following representative models to report our accuracies. Note that the ensemble includes "none of the above category" optimization as well.
\begin{table}[h]
\centering
\caption{Accuracy Results for Various Models}
\label{my-label}
\begin{tabular}{lll}
                                                    & \multicolumn{2}{l}{\textbf{Validation Accuracy}} \\
\textbf{Models}                                     & 25 Epochs             & 50 Epochs                \\
MNIST Model Baseline                                & 67.51\%               & 68.93\%                  \\
64 Capsule Layers                                   & 60.54\%               & 64.67\%                  \\
4-Model Ensemble (4 Ensemble)                       & 68.97\%               & 70.78\%                  \\
2-Convolution Layers (2 Conv)                       & 68.14\%               & 69.34\%                  \\
4 Ensemble + 2 Conv                                 & 70.34\%               & 71.50\%                  \\
7 Ensemble + 2 Conv                                 & 70.50\%               & \_\_\_\_\_\_             \\
4 Ensemble + 2 Conv + 0.0001 Reconstruction Scaling & 69.21\%               & \_\_\_\_\_\_             \\
Stack Additional Capsule Layer                      & 10.11\%               & \_\_\_\_\_\_            
\end{tabular}
\end{table}

We train for 50 epochs due to both limitation in resources and the observation that validation accuracy plateaus around that epoch number (as shown in the figures below). \\
\begin{figure}
 \centering
\includegraphics[scale=0.6]{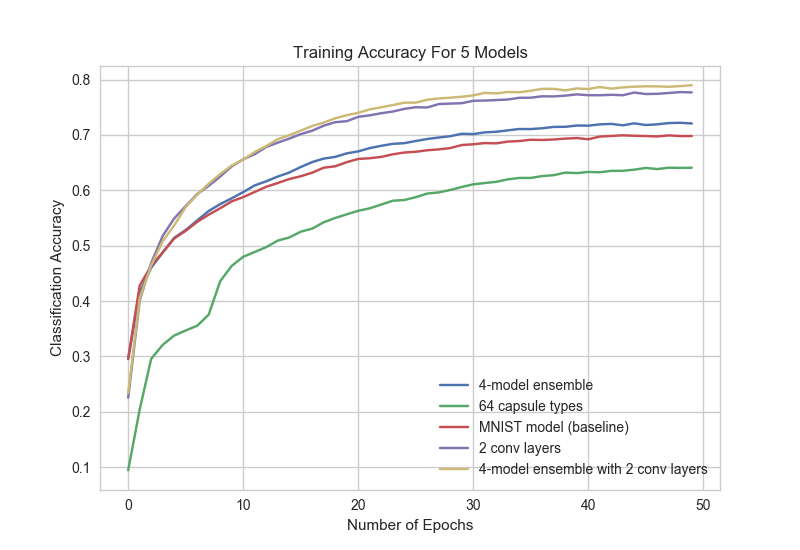}
\end{figure}
\begin{figure}
\centering
\includegraphics[scale=0.6]{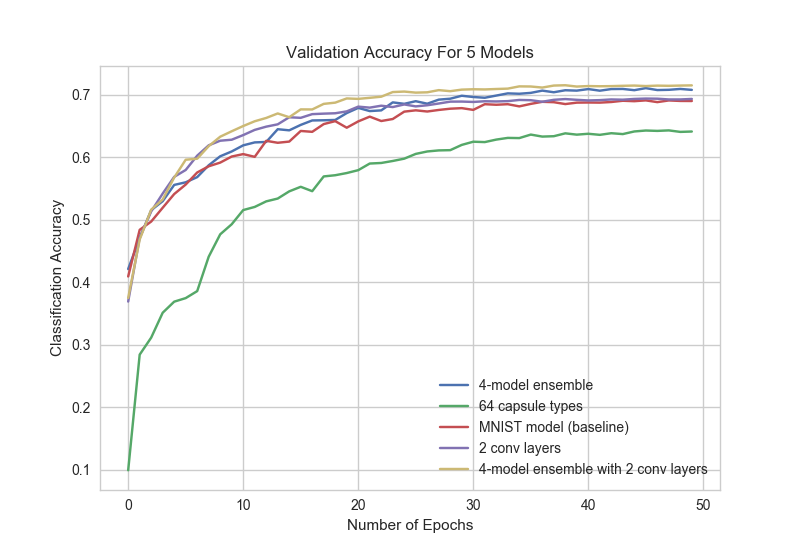}
\end{figure}
Despite the success in several of the modifications, such as adding a convolution layer and ensemble averaging, most of the modifications proved to be inferior than the baseline. Contrary to expectation, stacking an additional capsule layer and using our own activation function performed drastically lower than the baseline model. Lowering reconstruction scaling and increasing the number of capsule types also underperforms than expected. \\
\\
The real and reconstruction images for both MNIST and CIFAR10 on the baseline model is shown below. The top half is the input, and the bottom half is the reconstructed images. As observed in the figure below, MNIST reconstruction exhibits clear structure and distinct features. By contrast, CIFAR10 reconstructions are blurry and lack distinct features for each class. We will provide a proposed explanation of this drastic difference in reconstruction iagmes in the section below, and provide a potential way to resolve it.

\includegraphics[scale=0.3]{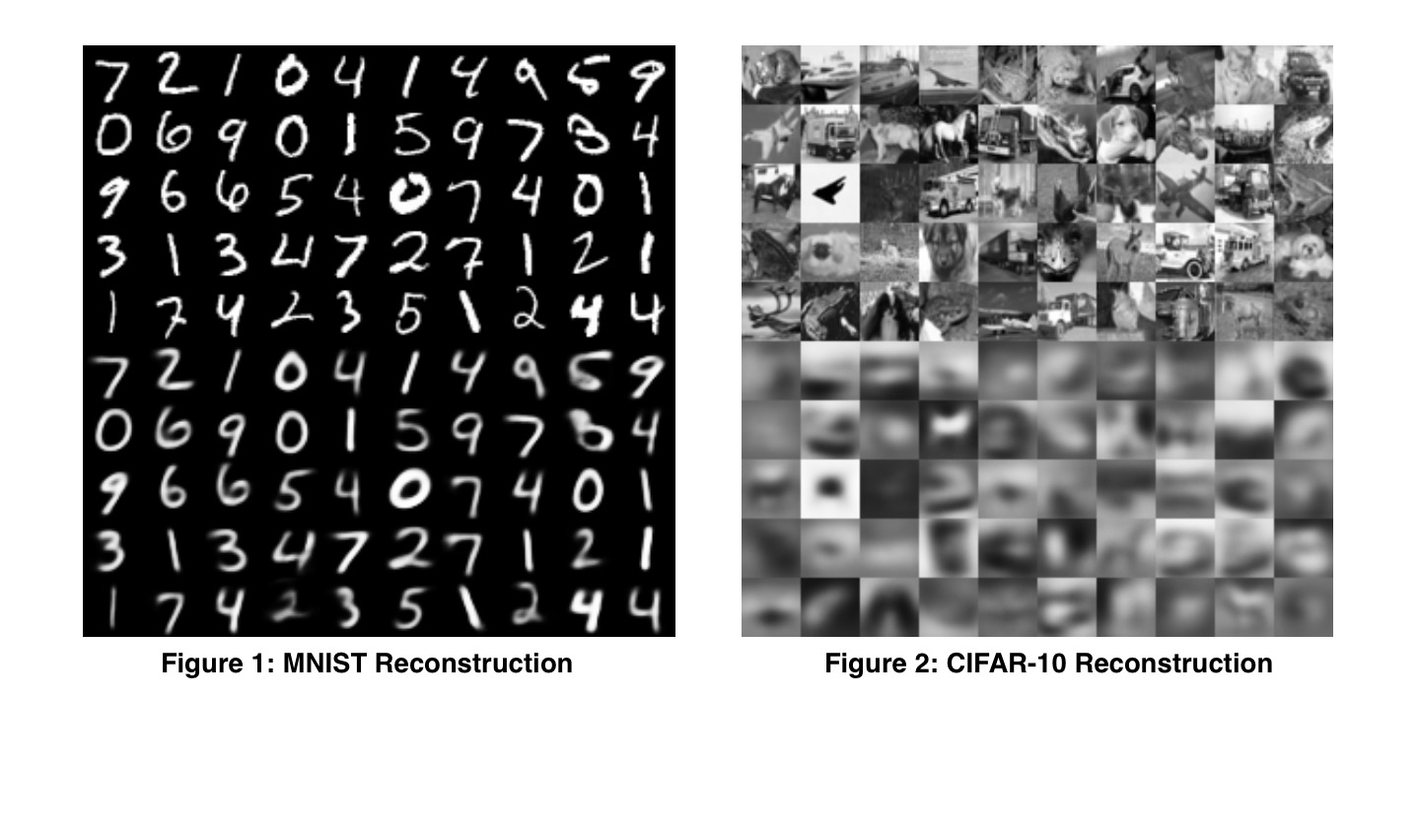}

\section{Discussion and Analysis}
\subsection{Model Comparisons}
Through experimentations, we find that adding a convolution layer increases validation accuracy by 0.41\%, and using a 4-model ensemble increases validation accuracy by 1.85\%, compared to the baseline model. The underlying reason for the successes of convolution and ensembling could be attributed to the need to represent complex features better in higher dimensionality. The most promising model out of all experimented is the model that utilizes 7-model ensemble with an additional convolution layer, surpassing its "4-ensemble 2-conv" network by 0.16\% as early as $25^{th}$ epoch.

\subsection{Summary}
Due to computational constraints, we compare the performance of different models at the 50 epoch mark, rather than training each model until absolute convergence. The baseline model in [1] uses no ensembling, one convolution layer and 32 types of capsules, achieving 68.93\% validation accuracy by 50 epochs. 

The best validation accuracy we reached is 71.550\% trained over 50 epochs, using a 4 ensemble model with 2 convolution layers. This represents a 2.57\% accuracy improvement over the baseline MNIST model introduced in the paper. This is a significant improvement over baseline, however our results fell short of the original paper's performance on CIFAR-10 using a 70-model ensemble and 64 capsules. 

Due to limits in computational resources, the combination of a 7-model ensemble and a 64-capsule network could not be tested. However, based on the performance of 2-model ensemble and 4-model ensemble, it is unlikely that the current best model in this paper will outperform Hinton's result by a significant amount.

\subsection{Reconstruction Loss}
The performance gap between MNIST and CIFAR10 may possibly be attributed to the reconstruction method. Unlike traditional neural nets, capsule networks' regularization technique attempts to minimize the difference between the reconstruction and true image. We also observe that capsule network is robust to affine transformations, which is a 2D transformation. This type of regularization works exceptionally well on 2-dimensional handwritten digits, where all of the transforms are either affine or rotational. However, to correctly classify a 3-dimensional object in the real world, viewpoint invariance -the ability to recognize objects regardless of viewing angles- is required. Unlike digits in MNIST, objects in CIFAR10 have more than one viewpoint for each class due to changes in viewpoint. Applying a 2-dimensional reconstruction regularization method on 3-dimensional data may cause inaccurate reconstruction regularization values, which may be a contributing factor to the subpar performance of capsule net on complex data. 
\subsection{Future Work}
A recently published paper in open review developed the idea of matrix capsules using EM routing[6], in which a 4 x 4 pose matrix is used to account for the relation between the object and viewer's position. As the viewpoint changes, the pose matrix is modified in a way such that the votes from different capsules will persist, thereby allowing capsule network with pose matrix to be viewpoint invariant. The inclusion of such a pose matrix is an interesting and promising direction of research in the future, as it seems to handle capsule network's shortcomings on complex data.

\newpage

\section*{References}



[1] Sabour, Sara, Nicholas Frosst, and Geoffrey E. Hinton. "Dynamic Routing Between Capsules." arXiv preprint arXiv:1710.09829 (2017).

[2] Krizhevsky, Alex, Ilya Sutskever, and Geoffrey E. Hinton. "Imagenet classification with deep convolutional neural networks." Advances in neural information processing systems. 2012.

[3] Hinton, Geoffrey E., Alex Krizhevsky, and Sida D. Wang. "Transforming auto-encoders." International Conference on Artificial Neural Networks. Springer, Berlin, Heidelberg, 2011.

[4] Krizhevsky, Alex, and Geoffrey Hinton. "Learning multiple layers of features from tiny images." (2009).

[5] Alex Krizhevsky.  "Learning Multiple Layers of Features from Tiny Images",  2009.

[6] Anon. Authors. "Matrix Capsules With EM routing" Paper under review, 2017

[7] Hinton, Geoffrey F. "A parallel computation that assigns canonical object-based frames of reference." Proceedings of the 7th international joint conference on Artificial intelligence-Volume 2. Morgan Kaufmann

\end{document}